%% file: camera-ready.tex
\documentclass[sigconf]{acmart}

\input{header.tex}

\begin{document}
\title{Emotion Recognition in Speech using \\ 
       Cross-Modal Transfer in the Wild} 
       
\input{acm-metadata.tex}
\input{author.tex}
\input{teaser.tex}
\input{abstract.tex}

\maketitle

\input{introduction}
\input{related}
\input{transfer}

\input{dataset}
\input{experiments}
\input{conclusion}

\input{acknowledgements}

\bibliographystyle{abbrv}
\bibliography{shortstrings,vgg_local,vgg_other,refs,mybib}

\end{document}

%% file: header.tex
\usepackage{booktabs} % For formal tables
\usepackage{dirtytalk}
\newcommand{\datasetName}{\textsc{EmoVoxCeleb}}

\newcommand\blfootnote[1]{%
  \begingroup
  \renewcommand\thefootnote{}\footnote{#1}%
  \addtocounter{footnote}{-1}%
  \endgroup
}

\acmArticle{4}
\acmPrice{15.00}

%% file: acm-metadata.tex
%  Index terms should be included as shown below.
%\begin{CCSXML}
%<ccs2012>
%<concept>
%<concept_id>10010147.10010257.10010293.10010294</concept_id>
%<concept_desc>Computing methodologies~Neural networks</concept_desc>
%<concept_significance>500</concept_significance>
%</concept>
%<concept>
%<concept_id>10010147.10010178.10010179.10010181</concept_id>
%<concept_desc>Computing methodologies~Discourse, dialogue and pragmatics</concept_desc>
%<concept_significance>300</concept_significance>
%</concept>
%</ccs2012>
%\end{CCSXML}
%
%\ccsdesc[500]{Computing methodologies~Neural networks}
%\ccsdesc[300]{Computing methodologies~Discourse, dialogue and pragmatics}

\keywords{Cross-modal transfer, speech emotion recognition}

\copyrightyear{2018} 
\acmYear{2018} 
\setcopyright{creativeCommons} 
%\setcopyright{rightsretained} 
\acmConference[MM '18]{2018 ACM Multimedia Conference}{October 22--26, 2018}{Seoul, Republic of Korea}
\acmBooktitle{2018 ACM Multimedia Conference (MM '18), October 22--26, 2018, Seoul, Republic of Korea}\acmDOI{10.1145/3240508.3240578}
\acmISBN{978-1-4503-5665-7/18/10}

% Add to arxiv version
%This work is licensed under a Creative Commons Attribution International 4.0 License.

%% file: author.tex
\author{Samuel Albanie*,  Arsha Nagrani*,  Andrea Vedaldi,  Andrew Zisserman \hspace{2em} \\
Visual Geometry Group, Department of Engineering Science, University of Oxford\\
{\tt\small \{albanie,arsha,vedaldi,az\}@robots.ox.ac.uk}}

%% file: teaser.tex
\begin{teaserfigure}
    \centering
    \includegraphics[trim={0 2cm 0 0},clip,width=0.9\textwidth]{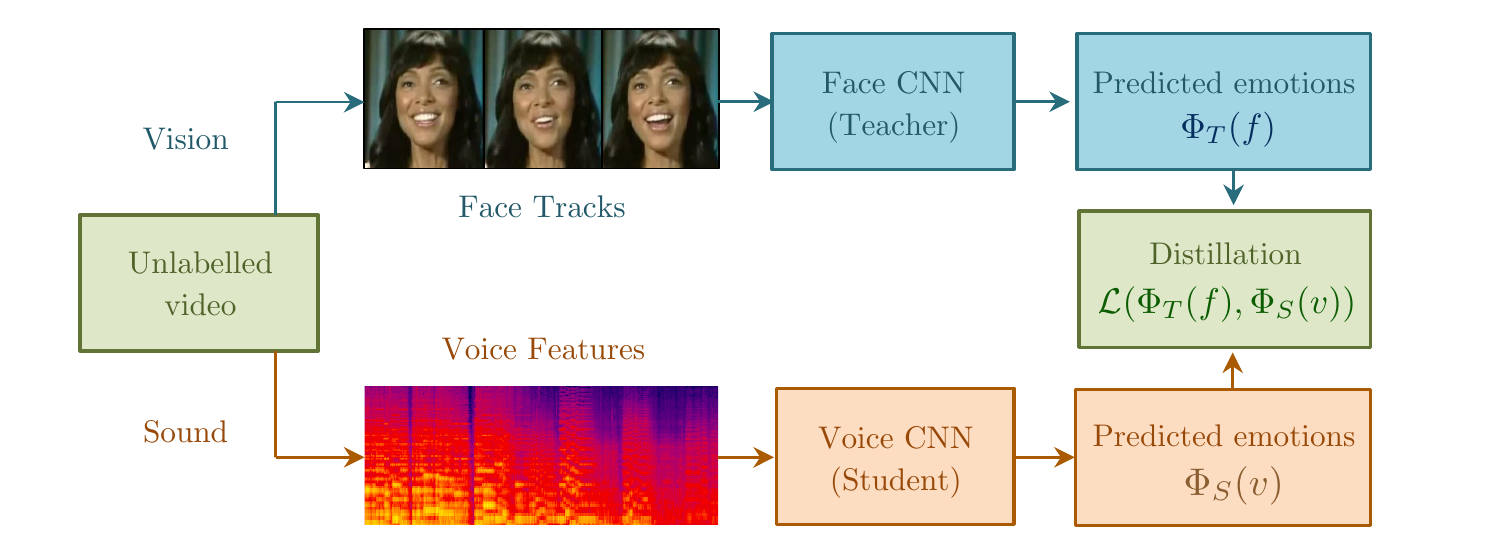}
    \caption{{\textbf Cross-modal transfer}:  A  CNN for speech emotion recognition (the student, $\Phi_S$) 
is trained by distilling the knowledge of a pretrained facial emotion recognition network (the teacher, $\Phi_T$) across unlabelled video. The student aims to exploit redundancy between the audio and visual signals $v$ and $f$ to learn embeddings, reducing dependence on labelled speech.}
    \vspace{1em}
\end{teaserfigure}

%% file: abstract.tex
\begin{abstract}
Obtaining large, human labelled speech datasets to train models for emotion recognition is a notoriously challenging task, hindered by annotation cost and label ambiguity. In this work, we consider the task of learning embeddings for speech classification without access to any form of labelled audio.  
We base our approach on a simple hypothesis: that the emotional content of speech correlates with the facial expression of the speaker.  By exploiting this relationship, we show that annotations of expression can be transferred from the visual domain (faces) to the speech domain (voices) through \textit{cross-modal distillation}.  We make the following contributions: 
(i) we develop a strong teacher network for facial emotion recognition
that achieves the state of the art on a standard benchmark;
(ii) we use the teacher to train a student, \textit{tabula rasa}, to learn representations (embeddings) for speech emotion recognition \textit{without access to labelled audio data}; and
(iii) we show that the speech emotion embedding can be used for speech emotion recognition on external benchmark datasets.
Code, models and data are available\footnote{\url{http://www.robots.ox.ac.uk/~vgg/research/cross-modal-emotions}}. 
\end{abstract}

%% file: introduction.tex
\blfootnote{*Equal contribution.}

\section{Introduction}

Despite recent advances in the field of speech emotion recognition, learning representations for ~\textit{natural} speech segments that can be used efficiently under noisy and unconstrained conditions still represents a significant challenge. Obtaining large, labelled human emotion datasets `in the wild' is hindered by a number of difficulties. First, since labelling naturalistic speech segments is extremely expensive, most datasets consist of elicited or acted speech. Second, as a consequence of the subjective nature of emotions, labelled datasets often suffer from low human annotator agreement, as well as the use of varied labelling schemes (i.e., dimensional or categorical) which can require careful alignment \cite{mariooryad2013analysis}. Finally, cost and time prohibitions often result in datasets with low speaker diversity, making it difficult to avoid speaker adaptation. Fully supervised techniques trained on such datasets hence often demonstrate high accuracy for only intra-corpus data, with a natural propensity to overfit~\cite{transfer2018Latif}.

In light of these challenges, we pose the following question: is it possible to learn a representation for emotional speech content for natural speech, from~\textit{unlabelled} audio-visual speech data, simply by transferring knowledge from the facial expression of the speaker? 

Given the recent emergence of large-scale  video datasets of human speech, it is possible to obtain examples of unlabelled human emotional speech at massive scales. Moreover, although it is challenging to assess the accuracy of emotion recognition models precisely, recent progress in computer vision has nevertheless enabled deep networks to learn to map faces to emotional labels in a manner that consistently matches a pool of human annotators \cite{Albanie16}. We show how to transfer this discriminative visual knowledge into an audio network using unlabelled video data as a bridge.  Our method is based on a simple hypothesis: that the emotional content of speech correlates with the facial expression of the speaker.

Our work is motivated by the following four factors. First, we would like to learn from a large,~\textit{unlabelled} collection of `talking faces' in videos as a source of free supervision, without the need for any manual annotation. Second, evidence suggests that this is a possible source of supervision that infants use as their visual and audio capabilities develop~\cite{grossmann2010development}. Newborns look longer at face-like stimuli and track them farther than non-face-like stimuli (Goren et al.~\cite{goren1975visual}; Johnson et al.~\cite{johnson1991newborns}), and combining these facial stimuli together with voices, detect information that later may allow for the discrimination and recognition of emotional expressions. Our third motivation is that we would like to be able to handle ambiguous emotions gracefully.  To this end, we seek to depart from annotation that relies on a single categorical label per segment, but instead incorporate a measure of uncertainty into the labelling scheme, building on prior work by~\cite{zhao2017learning} and ~\cite{han2017hard}.  Finally, accepting that the relationship between facial and vocal emotion will be a noisy one, we would like to make use of the remarkable ability of CNNs to learn effectively in the presence of label noise when provided with large volumes of training data \cite{rolnick2017deep, mahajan2018exploring}.

We make the following contributions: (i) we develop a strong model for
facial expression emotion recognition, achieving state of the art
performance on the FERPlus benchmark (section~\ref{sec:teacher}), (ii) we use this computer vision
model to label face emotions in the VoxCeleb~\cite{Nagrani17} video dataset -- this is
a large-scale dataset of emotion-unlabelled speaking face-tracks
obtained in the wild (section~\ref{sec:dataset}); (iii) we 
 transfer supervision \textit{across modalities} from faces to a speech, and then train a
speech emotion recognition model using speaking face-tracks (section~\ref{s:experiments});
 and,  (iv) we demonstrate that the resulting
speech model is capable of classifying emotion on two external
datasets (section~\ref{sec:external}).  A by-product of our method is that we obtain emotion
annotation for videos in the VoxCeleb dataset automatically using the facial expression model, which we release as the \datasetName{} dataset.

%% file: related.tex
\section{Related Work}\label{sec:related}

\textbf{Teacher-student methods}.
Teaching one model with another was popularised by
\cite{bucilua2006model} who trained a single model to match the
performance of an ensemble, in the context of model
compression. Effective supervision can be provided by the ``teacher''
in multiple ways: by training the ``student'' model to regress the
pre-softmax logits \cite{ba2014deep}, or by minimising cross entropy
between both models' probabilistic outputs \cite{li2014learning},
often through a high-temperature softmax that softens the predictions
of each model \cite{hinton2015distilling,crowley2017moonshine}. In
contrast to these methods which transfer supervision within the same
modality, \textit{cross-modal} distillation obtains supervision in one
modality and transfers it to another.  This approach was proposed for
RGB and depth paired data, and for RGB and flow paired data by
\cite{gupta2016cross}.  
More recent
work~\cite{aytar2016soundnet,arandjelovic2017look,aytar2017see,owens2016ambient}
has explored this concept by exploiting the correspondence between
synchronous audio and visual data in teacher-student style
architectures~\cite{aytar2016soundnet,aytar2017see}, or as a form of
\say{self-supervision}~\cite{arandjelovic2017look} where networks for
both modalities are learnt from scratch (an idea that was previously explored in the neuroscience community \cite{barlow1989unsupervised}).   Some works have also
examined cross-modal relationships between faces and voices in order
to learn identity
representations~\cite{Nagrani18a,nagrani2018learnable,kim2018learning}. Differently
from these works, our approach places an explicit reliance on the
correspondence between the facial and vocal {\em emotions} emitted by
a speaker during speech, discussed next.

\noindent \textbf{Links between facial and vocal emotion}. 
Our goal is to learn a representation that is aware of the emotional
content in speech \textit{prosody}, where prosody refers to the
extra-linguistic variations in speech (e.g.\ changes in pitch, tempo,
loudness, or intonation), by transferring such emotional knowledge
from face images extracted synchronously. For this to be possible, the
emotional content of speech must correlate with the facial expression
of the speaker. Thus in contrast to multimodal emotion recognition
systems which seek to make use of the \text{complementary} components
of the signal between facial expression and speech
\cite{busso2004analysis}, our goal is to perform cross-modal learning
by exploiting the redundancy of the signal that is \text{common} to
both modalities.  Fortunately, given their joint relevance to
communication, person perception, and behaviour more generally,
interactions between speech prosody and facial cues have been
intensively studied (Cvejic {\it et al.}~\cite{cvejic2010prosody};
Pell~\cite{pell2005prosody}; Swerts and Krahmer~\cite{swerts2008facial}).  The broad consensus of these works is
that during conversations, speech prosody is typically associated with
other social cues like facial expressions or body movements, with
facial expression being the most `privileged' or informative
stimulus~\cite{rigoulot2014emotion}.

\noindent \textbf{Deep learning for speech emotion recognition}.
Deep networks for emotional speech recognition either operate on hand-crafted acoustic features known to have a significant effect on speech prosody, (e.g.\  MFCCs, pitch, energy, ZCR, ...), or operate on raw audio with little processing, e.g.\  only the application of Fourier transforms~\cite{cummins2017image}. 
Those that use handcrafted features focus on global suprasegmental/prosodic features for emotion recognition, in which utterance level statistics are calculated. The main limitation of such global-level acoustic features is that they cannot describe the dynamic variation along an utterance~\cite{aldeneh2017using}. Vocal emotional expression is shaped to some extent by differences in the temporal structure of language and emotional cues are not equally salient throughout the speech signal~\cite{rigoulot2014emotion,kim2016emotion}.
In particular,  there is a well-documented propensity for speakers to elongate syllables located in word- or phrase-final positions~\cite{oller1973effect,pell2001influence}, and evidence that speakers vary their pitch in final positions to encode gradient acoustic cues that refer directly to their emotional state (Pell~\cite{pell2001influence}).  
We therefore opt for the second strategy, using minimally processed audio represented by magnitude
spectrograms directly as inputs to the network. Operating on these features can potentially improve performance ``in the wild'' where the encountered input can be unpredictable and diverse~\cite{kim2017deep}. By using CNNs with max pooling on spectrograms, we encourage the network to determine the emotionally salient regions of an utterance.

\input{tab-existing}

\noindent \textbf{Existing speech emotion datasets}.
Fully supervised deep learning techniques rely heavily on large-scale labelled datasets, which are tricky to obtain for emotional speech. Many methods rely on using actors~\cite{burkhardt2005database,martin2006enterface,liberman2002ldc,busso2008iemocap} (described below), and automated methods are few. Some video datasets are created using subtitle analysis \cite{dhall2012collecting}. In the facial expression domain, labels can be generated through reference events \cite{Albanie16}, however this is challenging to imitate for speech. A summary of popular existing datasets in given in Table~\ref{tab:datasets}. We highlight some common disadvantages of these datasets below, and contrast these with
the VoxCeleb dataset that is used in this paper:

\vspace{1mm}
\noindent (1) 
Most speech emotion datasets consist of elicited or acted speech, typically created in a recording studio, where actors read from written text. However, as~\cite{douglas2000new} points out, full-blown emotions very rarely appear in the real world and models trained on
acted speech rarely generalise to natural speech. Furthermore there
are physical emotional cues that are difficult to consciously mimic,
and only occur in natural speech. In contrast,  VoxCeleb
consists of interview videos from  YouTube, and so is more naturalistic.

\vspace{1mm}
\noindent (2) Studio recordings are also often extremely clean and do not suffer from `real world' noise artefacts. In contrast, videos in the VoxCeleb dataset are degraded with real world noise, consisting of background chatter, laughter, overlapping speech and room acoustics.  The videos also exhibit considerable variance in the quality of recording equipment and channel noise. 

\vspace{1mm}
\noindent (3)  For many existing datasets, cost and time prohibitions result in low speaker diversity, making it difficult to avoid speaker adaptation. Since our method does not require any emotion labels, we can train on VoxCeleb which is two orders of magnitude larger than existing public speech emotion datasets in the number of speakers.
\vspace{1mm}

Note that for any machine learning system that aims to perform emotion recognition using vision or speech, the ground truth emotional state of the speaker is typically unavailable.  To train and assess the performance of models, we must ultimately rely on the judgement of human annotators as a reasonable proxy for the true emotional state of a speaker.  Throughout this work we use the term \say{emotion recognition} to mean accurate prediction of this proxy. 

%% file: tab-existing.tex
\begin{table*}[ht]
\setlength{\tabcolsep}{0.7em} % for the horizontal padding
%\footnotesize
%\centering
\begin{tabular}{llllc}
\multicolumn{1}{l}{Corpus}&\multicolumn{1}{l}{Speakers}&\multicolumn{1}{l}{Naturalness} & \multicolumn{1}{l}{Labelling method}&\multicolumn{1}{c}{Audio-visual}  \\   \hline         
AIBO$\star$~\cite{batliner2004you}  &  51 & Natural  &  Manual & Audio only \\
EMODB~\cite{burkhardt2005database}  & 10  & Acted  & Manual &  Audio only \\
ENTERFACE~\cite{martin2006enterface}  & 43  & Acted  &  Manual & $\checkmark$ \\
LDC~\cite{liberman2002ldc}  & 7  & Acted  &  Manual & Audio only \\
IEMOCAP~\cite{busso2008iemocap}  & 10  & Both$\dagger$  & Manual & $\checkmark$ \\
AFEW 6.0$\spadesuit$~\cite{dhall2012collecting}  & unknown$^+$ & Acted  &  Subtitle Analysis & $\checkmark$ \\
RML  &  8 & Acted  &  Manual & $\checkmark$ \\
{\tt \datasetName{}}  &  \textbf{1,251} &  \textbf{Natural} & \textbf{Expression Analysis} &$\checkmark$ \\  \hline         
\vspace{5pt}   
\end{tabular}
\caption{Comparison to existing public domain speech emotion datasets. $\dagger$ contains both improvised and scripted speech. $\star$ contains only emotional
speech of children. $\spadesuit$ has not been commonly used for audio only classification, but is popular for audio-visual fusion methods. $^+$ identity labels are not provided. }
\label{tab:datasets}
\end{table*}

%% file: transfer.tex
\section{Cross Modal Transfer}\label{sec:distillation}

The objective of this work is to learn useful representations for emotion speech recognition, without access to labelled speech data. Our approach, inspired by the method of cross modal distillation~\cite{gupta2016cross}, is to tackle this problem by exploiting readily available annotated data in the visual domain.

Under the formulation introduced in~\cite{gupta2016cross}, a \say{student} model operating on one input modality learns to reproduce the features of a \say{teacher} model, which has been trained for a given task while operating on a different input modality (for which labels are available). The key idea is that by using a sufficiently large dataset of modality paired inputs, the teacher can transfer task supervision to the student without the need for labelled data in the student's modality. Importantly, it is assumed that the paired inputs possess the same attributes with respect to the task of interest.

In this work, we propose to use the correspondence between the emotion expressed by the facial expression of a speaker and the emotion of the speech utterance produced synchronously.  Our approach relies on the assumption that there is some redundancy in the emotional content of the signal communicated through the concurrent expression and speech of a speaker.   To apply our method, we therefore require a large number of \textit{speaking face-tracks}, in which we have a known correspondence between the speech audio and the face depicted.  Fortunately, this can be acquired, automatically and at scale using the recently developed SyncNet~\cite{Chung16a}.  This method was used to generate the large-scale VoxCeleb dataset \cite{Nagrani17} for speaking face-tracks, which forms the basis of our study.

As discussed in Sec. \ref{sec:related}, there are several ways to \say{distill} the knowledge of the teacher to the student. While ~\cite{gupta2016cross} trained the student by regressing the intermediate representations at multiple layers in the teacher model, we found in practice that the approach introduced in \cite{hinton2015distilling} was most effective for our task.  Specifically, we used a cross entropy loss between the outputs of the networks after passing both both sets of predictions through a softmax function with temperature $T$ to produce a distribution of predictions:

\begin{equation}
p_i = \frac{\exp{(x_i/T)}}{\sum_j \exp{(x_j/T)}},  
\end{equation}

\noindent where $x_i$ denotes the logit associated with class $i$ and $p_i$ denotes the corresponding normalised prediction.  A higher temperature softmax produces a \say{softer} distribution over predictions. We experimented with several values of $T$ to facilitate training and found, similarly to \cite{hinton2015distilling}, that a temperature of $2$ was most effective.  We therefore use this temperature value in all reported experiments.  

\input{tab-ferplus}

\subsection{The Teacher\label{sec:teacher}}

This section describes how we obtain the teacher model which is responsible for classifying facial emotion in videos.

\noindent \textbf{Frame-level Emotion Classifier}. To construct a strong teacher network (which is tasked with performing emotion recognition from \textit{face images}), training is performed in multiple stages. We base our teacher model on the recently introduced Squeeze-and-Excitation architecture \cite{Hu18} (the ResNet-$50$ variant).  The network is first pretrained on the large-scale VGG-Face$2$ dataset \cite{Cao18} (${\approx3.3}$~million faces) for the task of identity verification.  The resulting model is then finetuned on the \textit{FERplus} dataset \cite{BarsoumICMI2016} for emotion recognition. This dataset comprises the images from the original FER dataset (${\approx35}$k~images) \cite{goodfellow2013challenges} together with a more extensive set of annotations ($10$ human annotators per image).  The emotions labelled in the dataset are: \textit{neutral}, \textit{happiness}, \textit{surprise}, \textit{sadness}, \textit{anger}, \textit{disgust}, \textit{fear} and \textit{contempt}.  Rather than training the teacher to predict a single correct emotion for each face, we instead require it to match the \textit{distribution} of annotator labels. Specifically, we train the network to match the distribution of annotator responses with a cross entropy loss:
\begin{equation}
\mathcal{L} = - \sum_n \sum_i p_i^{(n)} \log q_i^{(n)},
\end{equation}
where $p_i^{(n)}$ represents the probability of annotation $n$ taking emotion label $i$, averaged over annotators, and $q_i^{(n)}$ denotes the corresponding network prediction.

During training, we follow the data augmentation scheme comprising affine distortions of the input images introduced in \cite{yu2015image} to encourage robustness to variations in pose.  To verify the utility of the resulting model, we evaluate on the FERPlus benchmark,  following the test protocol defined in \cite{BarsoumICMI2016},  and report the results in Table~\ref{tab:FERplus}. To the best of our knowledge, our model represents the current state of the art on this benchmark.

\begin{figure}[]
\centering
\includegraphics[width=1\linewidth]{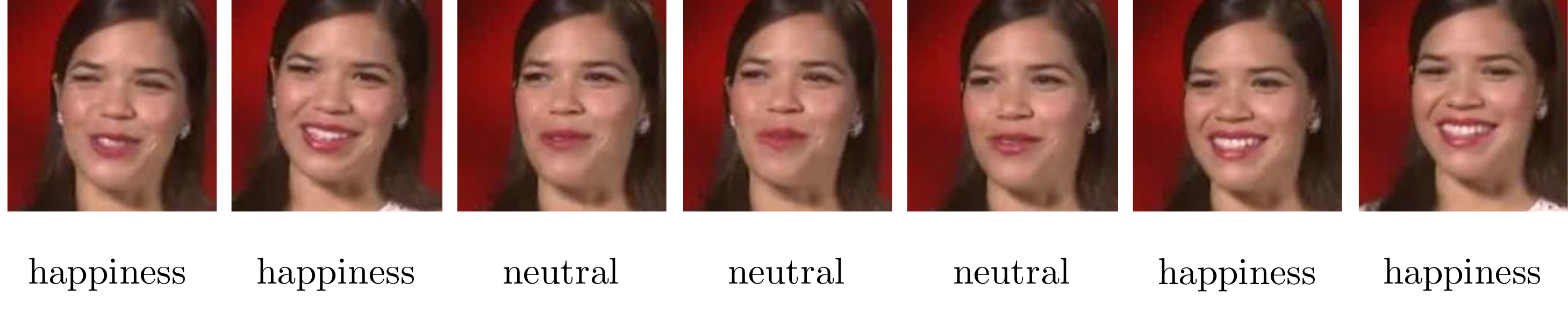}
   \caption{An example set of frames accompanying a single speech segment in the VoxCeleb dataset illustrating the \textit{neutral transition-face} phenomenon exhibited by many face tracks:  the facial expression of the speaker, as predicted by the static image-based face classifier often takes a `neutral' label while transitioning between certain phonemes.}
\label{fig:exfacetrack}
\end{figure}

\noindent \textbf{From Frames to Face-tracks}. Since a single speech segment typically spans many frames, we require labels at a face-track level in order to transfer knowledge from the face domain to the speech domain.  To address the fact that our classifier has been trained on individual images, not with ~\textit{face-tracks}, we take the simplest approach of considering a single face-track as a set of individual frames. A natural consequence of using still frames extracted from video, however, is that the emotion of the speaker is not captured with equal intensity in every frame.  Even in the context of a highly emotional speech segment, many of the frames that correspond to transitions between utterances exhibit a less pronounced facial expression, and are therefore often labelled as `neutral' (see Figure~\ref{fig:exfacetrack} for an example track). One approach that has been proposed to address this issue is to utilise a single frame or a subset of frames known as~\textit{peak frames}, which best represent the emotional content of the face-track~\cite{zhalehpour2016multimodal,poria2015towards}. The goal of this approach is to select the frames for which the dominant emotional expression is at its apex. It is difficult to determine which frames are the key frames, however, while ~\cite{poria2015towards} select these frames manually, ~\cite{zhalehpour2016multimodal} add an extra training step which measures the `distance' of the expressive face from the subspace of neutral facial expressions. This method also relies on the implicit assumption that all facial parts reach the peak point at the same time.

We adopt a simple approximation to peak frame selection by representing each track by the maximum response of each emotion across the frames in the track, an approach that we found to work well in practice.  We note that prior work has also found simple average pooling strategies over frame-level predictions~\cite{bargal2016emotion,hu2017learning} to be effective (we found average pooling to be slightly inferior, though not dramatically different in performance). To verify that max-pooling represents a reasonable temporal aggregation strategy, we applied the trained SENet Teacher network to the individual frames of the AFEW $6.0$ dataset, which formed the basis of the $2016$ Emotion Recognition in the Wild (EmotiW) competition~\cite{dhall2016emotiw}. Since our objective here is not to achieve the best performance by specialising for this particular dataset (but rather to validate the aggregation strategy for predicting tracks), we did not fine-tune the parameters of the teacher network for this task.  Instead, we applied our network directly to the default face crops provided by the challenge organisers and aggregated the emotional responses over each video clip using max pooling. We then treat the predictions as $8$-dimensional embeddings and use the AFEW training set to fit a single affine transformation (linear transformation plus bias), followed by a softmax, allowing us to account for the slightly different emotion categorisation (AFEW does not include a \textit{contempt} label).  By evaluating the resulting re-weighted predictions on the validation set we obtained an accuracy of $49.3\%$ for the $7$-way classification task, strongly outperforming the baseline of $38.81\%$ released by the challenge organisers. 

\subsection{The Student}

The student model, which is tasked with performing emotion recognition \textit{from voices}, is based on the VGG-M architecture \cite{Chatfield14} (with the addition of batch normalization). This model has proven effective for speech classification tasks in prior work \cite{Nagrani17}, and provides a good trade-off between computational cost and performance. The architectural details of the model are described in section~\ref{sec:implement}.

\subsection{Time-scale of transfer} 
The time-scale of transfer determines the length of the audio segments that are fed into the student network for transferring the logits from face to voice. Determining  the optimal length of audio segment for which emotion is discernable is still an open question. Ideally, we would like to learn only features related to speech~\textit{prosody} and not the lexical content of speech, and hence we do not want to feed in audio segments that contain entire sentences to the student network. We also do not want segments that are too short, as this creates the risk of capturing largely neutral audio segments. 
%that the audio segments are not long enough that the network learns to fit to the actual words being said. 
Rigoulot, 2014~\cite{rigoulot2014emotion} studied the time course for recognising vocally expressed emotions on human participants, and found that while some emotions were more quickly recognised than others (fear as opposed to happiness or disgust), after four seconds of speech emotions were usually classified correctly.
%most emotions were recognised within a $400-1200$ms time window. 
We therefore opt for a four second speech segment input. Where the entire utterance is shorter than four seconds, we use zero padding to obtain an input of the required length.

%% file: tab-ferplus.tex
\begin{table}
\setlength{\tabcolsep}{0.7em} % for the horizontal padding     
\centering
%\footnotesize
\begin{tabular}{cc}
\multicolumn{1}{c}{Method} & \multicolumn{1}{c}{Accuracy (\texttt{PrivateTest})} \\   \toprule         
PLD \cite{BarsoumICMI2016} & 85.1 $\pm 0.5\%$ \\
CEL \cite{BarsoumICMI2016} & 84.6 $\pm 0.4\%$ \\
ResNet+VGG$\dagger$ \cite{huang2017combining} & 87.4 \\
\hline
SENet Teacher (Ours) & 88.8  $\pm 0.3\%$ \\
% Trials (for reference)
% ---------
% PLD Trials: 85.4, 84.7, 85.3 - mean, std: (85.1, 0.5)
% CEL Trials: 85.0, 84.6, 84.3 - mean, std: (84.6, 0.4)
% SENet Teacher Trials: 88.6, 88.7, 89.1 - mean, std: ()
% Model trained on dataset containing additional ambiguous labels scores 86.50

\end{tabular}
\normalsize
\vspace{5pt}
\caption{Comparison on the FERplus facial expression benchmark.  $\dagger$ denotes performance of model ensemble.  Where available, the mean and std.\  is reported over three repeats. The 
SENet Teacher model is described in  Sec.~\ref{sec:teacher}.}
\label{tab:FERplus}
\end{table}

%% file: dataset.tex
\section{\datasetName{} Dataset}
\label{sec:dataset}

\begin{figure}[]
\centering
 \includegraphics[trim={1cm 1cm 1cm 1cm},clip,width=1\linewidth]{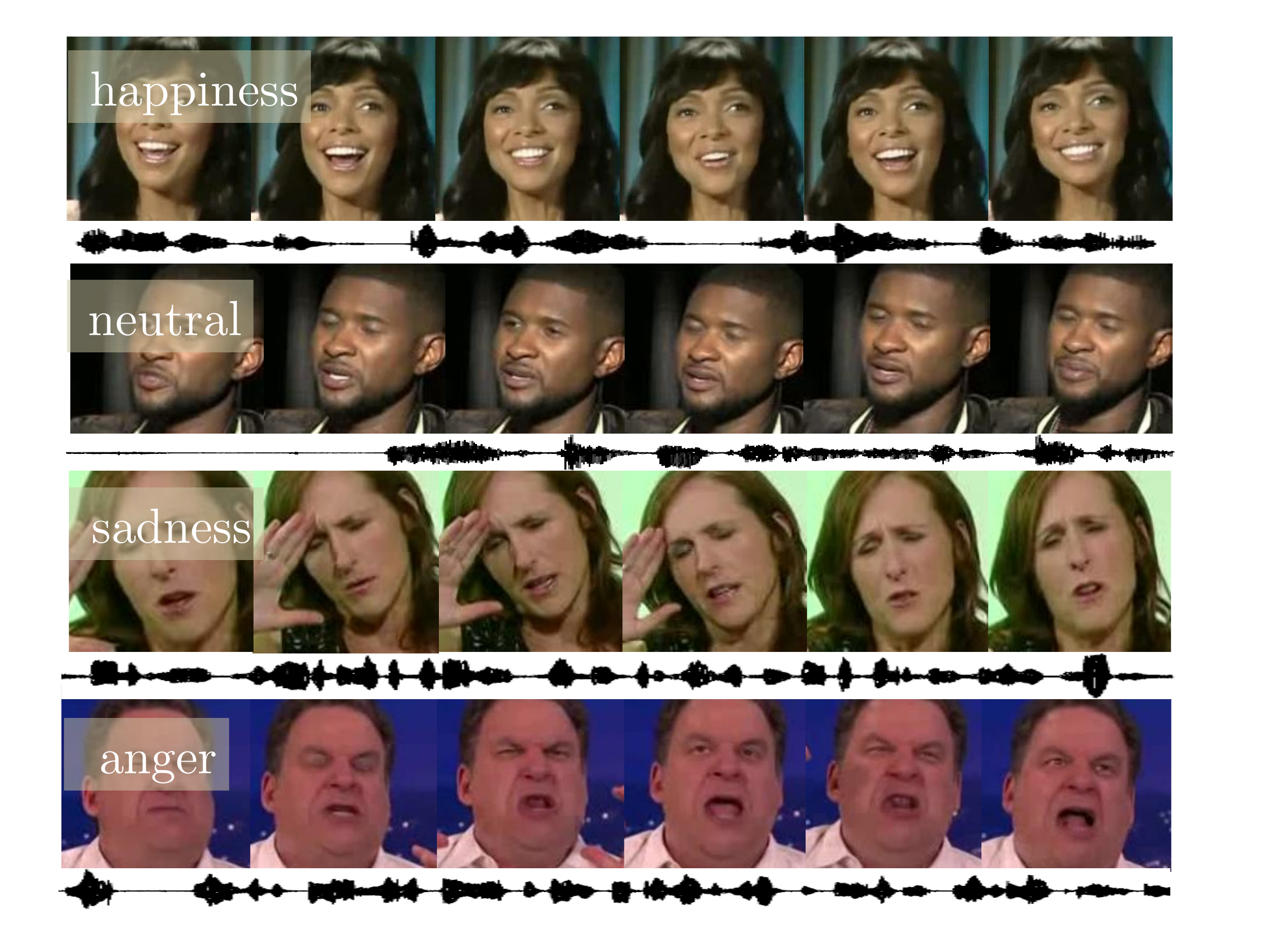}
   \caption{\small{Examples of emotions in the~\datasetName{} dataset.  We rely on the facial expression of the speaker to provide clues about the emotional content of their speech.}}
\label{fig:facetrack}
\end{figure}

We apply our teacher-student framework on the VoxCeleb \cite{Nagrani17} dataset, a collection of \textit{speaking face-tracks}, or contiguous groupings of talking face detections from video. The videos in the VoxCeleb dataset are interview videos of 1,251 celebrities uploaded to YouTube, with over 100,000 utterances (speech segments). The speakers span a wide range of different ages, nationalities, professions and accents. The dataset is roughly gender balanced. The audio segments also contain speech in different languages.
While the identities of the speakers are available, the dataset has~\textit{no emotion labels}, and the student model must therefore learn to reason about emotions entirely by transferring knowledge from the face network.  The identity labels allow us to partition the dataset into three splits: \texttt{Train}, \texttt{Heard-Val} and \texttt{Unheard-Val}.  The \texttt{Heard-Val} split contains 
%a mixture of identities which (ensuring that no speech segments overlap with the training set),
held out speech segments from the same identities in the training set, while the \texttt{Unheard-Val} split contains identities that are disjoint from the other splits\footnote{The \texttt{Unheard-Val} split directly corresponds to the \texttt{Test (US-UH)} set defined in \cite{nagrani2018learnable}.}.  Validating on unheard identities allows us to ascertain whether the student model is exploiting identity as a bias to better match the predictions of the teacher model.  The identity labels may also prove useful for researchers tackling other tasks, for example evaluating the effect of emotional speech on speaker verification, as done by~\cite{parthasarathy2017study}.  The total size of each partition is given in Table \ref{tab:stats}.

\setlength{\tabcolsep}{0.7em} % for the horizontal padding     
\begin{table}
\centering

\begin{tabular}{rccc}
\multicolumn{1}{c}{}&\multicolumn{1}{c}{\texttt{Train}}&\multicolumn{1}{c}{\texttt{Heard-Val}}&\multicolumn{1}{c}{\texttt{Unheard-Val}} \\   \hline         
\# speaking face-tracks  & 118.5k  & 4.5k & 30.5k    \\

\end{tabular}
\vspace{2ex}
\caption{ The distribution of speaking face-tracks in the \datasetName{} dataset. The \texttt{Heard-Val} set contains identities that are present in \texttt{Train}, while the identities in \texttt{Unheard-Val} are disjoint from \texttt{Train}.\label{tab:stats}}
\end{table}

By applying the teacher model to the frames of the VoxCeleb dataset as
described in section~\ref{sec:teacher}, we automatically obtain
emotion labels for the face-tracks and the speech segments. These
labels take the form of a predicted distribution over eight emotional
states that were used to train the teacher model: \textit{neutral},
\textit{happiness}, \textit{surprise}, \textit{sadness},
\textit{anger}, \textit{disgust}, \textit{fear} and \textit{contempt}.
These frame-level predictions can then be directly mapped to
synchronous speech segments by aggregating the individual prediction
distributions into a single eight-dimensional vector for each speech
segment. For all experiments we perform this aggregation by
max-pooling across frames.  However,
since the best way to perform this aggregation remains an open topic
of research, we release the frame level predictions of the model as
part of the dataset annotation.  The result is a large-scale
audio-visual dataset of human emotion, which we call the
~\datasetName{} dataset.  As a consequence of the automated labelling
technique, it is reasonable to expect that the noise associated with
the labelling will be higher than for a manually annotated dataset.
We validate our labelling approach by demonstrating
quantitatively that the labels can be used to learn useful speech
emotion recognition models (Sec. \ref{sec:external}).  Face-track visualisations can be seen in
Figure~\ref{fig:facetrack}, and audio examples are available
online\footnote{\url{http://www.robots.ox.ac.uk/~vgg/research/cross-modal-emotions}}.

\begin{figure}[]
\centering
 \includegraphics[width=1\linewidth]{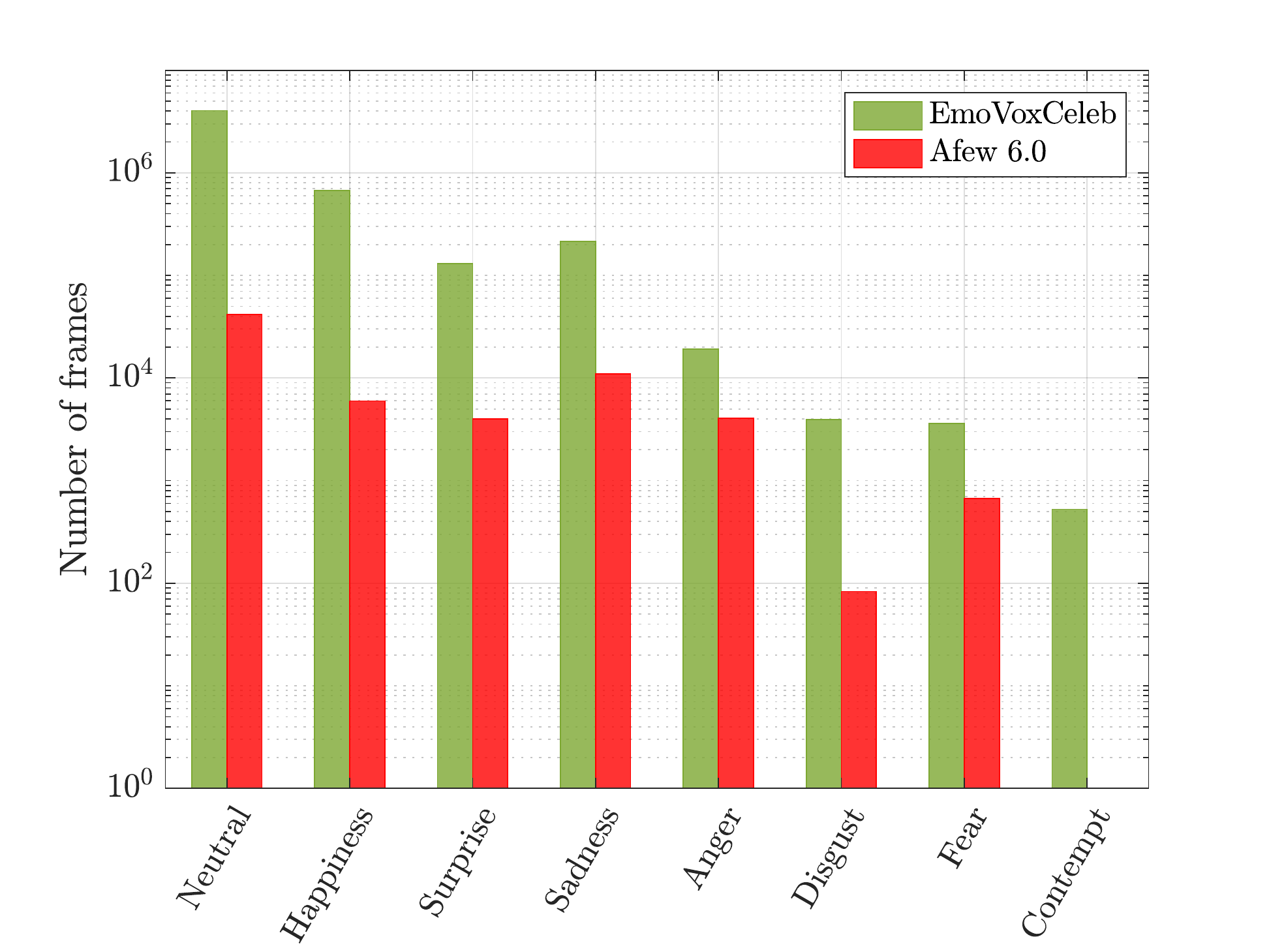}
   \caption{Distribution of frame-level emotions predicted by the SENet Teacher model for \datasetName{} (note that the y-axis uses a log-scale). For comparison, the distribution of predictions are also shown for the Afew $6.0$ dataset. }
\label{fig:distTracks}
\end{figure}

\noindent \textbf{Distribution of emotions}. As noted above, each frame of the dataset is annotated with a distribution of predictions. To gain an estimate of the distribution of emotional content in \datasetName, we plot a histogram of the \textit{dominant} emotion (the label with the strongest prediction score by the teacher model) for each extracted frame of the dataset, shown in Figure \ref{fig:distTracks}.  While we see that the dataset is heavily skewed towards a small number of emotions (particularly neutral, as discussed in Sec. \ref{sec:distillation}), we note that it still contains some diversity of emotion. For comparison, we also illustrate the distribution of emotional responses of the teacher model on `Afew $6.0$' \cite{dhall2012collecting}, an emotion recognition benchmark.  The Afew dataset was collected by selecting scenes in movies for which the subtitles contain highly emotive content.  We see the distribution of labels is significantly more balanced but still exhibits a similar overall trend to \datasetName. Since this dataset has been actively sampled to contain good diversity of emotion, we conclude that the coverage of emotions in \datasetName{} may still prove useful, given that no such active sampling was performed.  
We note that Afew does not contain segments directly labelled with the
\textit{contempt} emotion, so we would therefore not expect there to
be frames for which this is the predicted emotion.
It is also worth noting
that certain emotions are rare in our dataset. Disgust, fear and
contempt are not commonly exhibited during natural speech,
particularly in interviews and are therefore rare in the predicted
distribution.

\noindent \textbf{Data Format.}
As mentioned above, we provide logits (the pre-softmax predictions of the teacher network) at a frame level which can be used to directly produce labels at an utterance level (using max-pooling as aggregation).  The frames are extracted from the face tracks at an interval of $0.24$ seconds, resulting in a total of approximately $5$ million annotated individual frames.

%% file: experiments.tex
\section{Experiments}\label{s:experiments}

To investigate the central hypothesis of this paper, namely that it is possible to supervise a speech emotion recognition model with a model trained to detect emotion in faces, we proceed in two stages.  First, as discussed in Sec. \ref{sec:dataset}, we compute the predictions of the SENet Teacher model on the frames extracted from the VoxCeleb dataset.  The process of distillation is then performed by randomly sampling segments of speech, each four seconds in duration, from the training partition of this dataset.  While a fixed segment duration is not required by our method (the student architecture can process variable-length clips by dynamically modifying its pooling layer), it leads to substantial gains in efficiency by allowing us to batch clips together.  We experimented with sampling speech segments in a manner that balanced the number of utterance level emotions seen by the student during training.  However, in practice, we found that it did not have a significant effect on the quality of the learned student network and therefore, for simplicity, we train the student without biasing the segment sampling procedure.  

For each segment, we require the student to match the response of the teacher network on the facial expressions of the speaker that occurred \textit{during the speech segment}.  In more detail, the responses of the teacher on each frame are aggregated through max-pooling to produce a single $8$-dimensional vector per segment.  As discussed in Section~\ref{sec:distillation}, both the teacher and student predictions are passed through a softmax layer before computing a cross entropy loss. Similarly to \cite{hinton2015distilling}, we set the temperature of both the teacher and student softmax layers to $2$ to better capture the confidences of the teacher's predictions.  We also experimented with regressing the pre-softmax logits of the teacher directly with an Euclidean loss (as done in \cite{ba2014deep}), however, in practice this approach did not perform as well, so we use cross entropy for all experiments.  As with the predictions made by the teacher, the distribution of predictions made by the student are dominated by the neutral class so the useful signal is primarily encoded through the relative soft weightings of each emotion that was learned during the distillation process. The student achieves a mean ROC AUC of $0.69$ over the teacher-predicted emotions present in the unheard identities (these include all emotions except disgust, fear and contempt) and a mean ROC AUC of $0.71$ on validation set of heard identities on the same emotions.

\subsection{Implementation Details} \label{sec:implement}
The student network is based on the VGGVox network architecture described in~\cite{Nagrani17}, which has been shown to work well on spectrograms, albeit for the task of speaker verification. The model is based on the lightweight VGG-M architecture, however the fully connected {\em fc6} layer of dimension $9 \times n$ (support in both dimensions) is replaced by two layers -- a fully connected layer of $9 \times 1$ (support in the frequency domain) and an average pool layer with support $1 \times n$, where $n$ depends on the length of the input speech segment
(for example for a 4 second segment, $n=11$).  
This allows the network to achieve some temporal invariance, 
and at the same time keeps the output dimensions the same as those of the original fully connected layer. 
\begin{table}[ht]
\scriptsize
\centering
\begin{tabular}{| c | c |  c | c | c | c | }
  \hline
  \textbf{Layer}  &  \textbf{Support} & \textbf{Filt dim.} & \textbf{\# filts.} 
  & \textbf{Stride} & \textbf{Data size} \\ \hline 

  conv1 & 7$\times$7 & 1  & 96 & 2$\times$2 & 254$\times$198 \\ \hline
  mpool1 & 3$\times$3 & -  & - & 2$\times$2 & 126$\times$99\\ \hline 
  conv2 & 5$\times$5 & 96  & 256 & 2$\times$2 & 62$\times$49 \\ \hline
  mpool2 & 3$\times$3 & -  & - & 2$\times$2 & 30$\times$24 \\ \hline   
  conv3 & 3$\times$3 & 256  & 256 & 1$\times$1 & 30$\times$24 \\ \hline
  conv4 & 3$\times$3 & 256  & 256 & 1$\times$1 & 30$\times$24 \\ \hline
  conv5 & 3$\times$3 & 256  & 256 & 1$\times$1 & 30$\times$24 \\ \hline
  {\bf mpool5} & 5$\times$3 & -  & - & 3$\times$2 & 9$\times$11 \\ \hline   
  {\bf fc6} & 9$\times$1 & 256  & 4096 & 1$\times$1 & 1$\times$11 \\ \hline
  {\bf apool6} & 1$\times n$ & -  & - & 1$\times$1 & 1$\times$1 \\ \hline
  fc7   & 1$\times$1 & 4096  & 1024 & 1$\times$1 & 1$\times$1 \\ \hline
  fc8   & 1$\times$1 & 1024  & 1251 & 1$\times$1 & 1$\times$1 \\ \hline

\end{tabular} 
\normalsize
\vspace{10pt}
\caption{The CNN architecture for the student network. 
The data size up until {\it fc6} is depicted for a 4-second input,
but the network is able to accept inputs of variable lengths. 
Batchnorm layers are present after every conv layer.
}
\label{table:convnet}
\vspace{-18pt}
\end{table}
The input to the teacher image is an RGB image, cropped from the source frame to include only the face region (we use the face detections provided by the VoxCeleb dataset) resized to $224 \times 224$, followed by mean subtraction. The input to the student network is a short-term amplitude spectrogram, extracted from four seconds of raw audio using a Hamming window of width 25ms and step (hop) 10ms, giving spectrograms of size $512 \times 400$. At train-time, the four second segment of audio is chosen randomly from the entire speaking face-track, providing an effective form of data augmentation. Besides performing mean and variance normalisation on every frequency bin of the spectrogram, no other speech-specific processing is performed, e.g. silence removal, noise filtering, etc. (following the approach outlined in~\cite{Nagrani17}). While randomly changing the speed of audio segments can be useful as a form of augmentation for speaker verification~\cite{Nagrani17}, we do no such augmentation here since changes in pitch may have a significant impact on the perceived emotional content of the speech.

\noindent \textbf{Training Details.} The network is trained for 50 epochs (one epoch corresponds to approximately one full pass over the training data where a speech segment has been sampled from each video) using SGD with momentum (set to $0.9$) and weight decay (set to $0.0005$). The learning rate is initially set to $1E-4$, and decays logarithmically to $1E-5$ over the full learning schedule. The student model is trained from scratch, using Gaussian-initialised weights. We monitor progress on the validation set of unheard identities, and select the final model to be the one that minimises our learning objective on this validation set. 

\subsection{Results on external datasets\label{sec:external}}

To evaluate the quality of the audio features learned by the student model, we perform experiments on two benchmark speech emotion datasets.

\noindent\textbf{RML:} The RML emotion dataset is an acted dataset containing 720 audiovisual emotional expression samples with categorical labels: \textit{anger, disgust, fear, happiness, sadness and surprise}. This database is language and cultural background independent. The video samples were collected from eight human subjects, speaking six different languages (English, Mandarin, Urdu, Punjabi, Persian, Italian). To further increase diversity, different accents of English and Chinese were also included.

\noindent\textbf{eNTERFACE~\cite{martin2006enterface}:} The eNTERFACE dataset is an acted dataset (in English) recorded in a studio. Forty-two subjects of fourteen nationalities were asked to listen to six successive short stories, each of which was designed to elicit a particular emotion. The emotions present are identical to those found in the RML dataset. 

Both external datasets consist of acted speech, and are labelled by human annotators. Since the external datasets are obtained in a single recording studio, they are also relatively clean, in contrast to the noisy segments in \datasetName. We choose the RML dataset for evaluation specifically to assess whether our embeddings can generalise to multilingual speech.  Both datasets are class-balanced. 

\begin{table}[h]
\footnotesize
\centering
\begin{tabular}{lcccc}
\multicolumn{1}{c}{Method }&\multicolumn{2}{c}{RML }&\multicolumn{2}{c}{eNTERFACE}\\
\hline
  & Modality &  Acc. & Modality &  Acc. \\
\hline         
Random  & A &  16.7  & A &  16.7    \\  
Student & A & $49.7 \pm 5.4$ & A & $34.3 \pm 4.0$ \\  
Teacher & V & $72.6 \pm 3.9$ & V &  $48.3 \pm 4.9$ \\
Noroozi et al.~\cite{noroozi2017audio} & A & 65.3  & A & 47.1  \\

\end{tabular}
\normalsize
\caption{Comparison of method accuracy on RML and eNTERFACE using the evaluation protocol of \cite{noroozi2017audio}. Where available, the mean $\pm$ std.\  is reported. \label{tab:results}}
\end{table}
\vspace{-0.7cm}
 
We do not evaluate the predictions of the student directly, for two reasons: first, the set of emotions used to train the student differ from those of the evaluation test set, and second, while the predictions of the student carry useful signal, they skew towards neutral as a result of the training distribution.  We therefore treat the predictions as $8$-dimensional embeddings and adopt the strategy introduced in Sec. \ref{sec:teacher} of learning a map from the set of embeddings to the set of target emotions, allowing the classifier to re-weight each emotion prediction using the class confidences produced by the student.  In more detail, for each dataset, we evaluate the quality of the student model embeddings by learning a single affine transformation (comprising a matrix multiply and a bias) followed by a softmax to map the $8$ predicted student emotions to the target labels of each dataset.  Although our model has been trained using segments of four seconds in length, its dynamic pooling layer allows it to process variable length segments.  We therefore use the full speech segment for evaluation.

To assess the student model, we compare against the following baselines: the expected performance at chance level by a random classifier;  and the performance of the teacher network, operating on the faces modality. We also compare with the recent work of~\cite{noroozi2017audio}, whose strongest speech classifier consisted of a random forest using a combination of $88$ audio features inc. MFCCs, Zero Crossings Density (ZCD), filter-bank energies (FBE) and other pitch/intensity-related components. We report performance using $10$-fold cross validation (to allow comparison with~\cite{noroozi2017audio}) in Table \ref{tab:results}.  While it falls short of the performance of the teacher, we see that the student model performs significantly better than chance. These results indicate that, while challenging, transferring supervision from the facial domain to the speech domain is indeed possible.  
Moreover, we note that the conditions of the evaluation datasets
differ significantly from  those on which the student network was trained.
We discuss this domain transfer problem for emotional speech in the following section.

\begin{figure}[ht]
\centering
\includegraphics[width=0.24\textwidth]{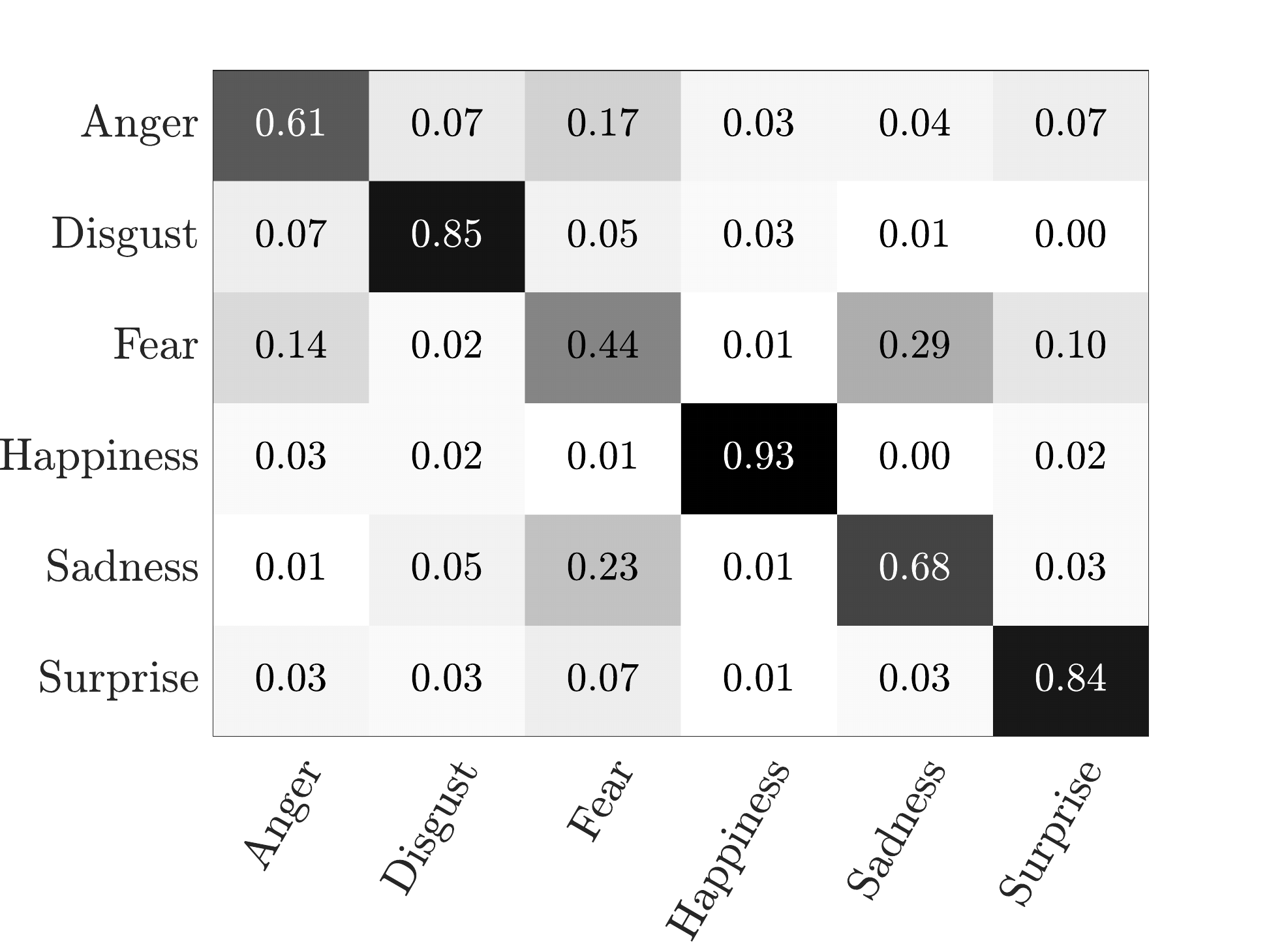}%
\includegraphics[width=0.24\textwidth]{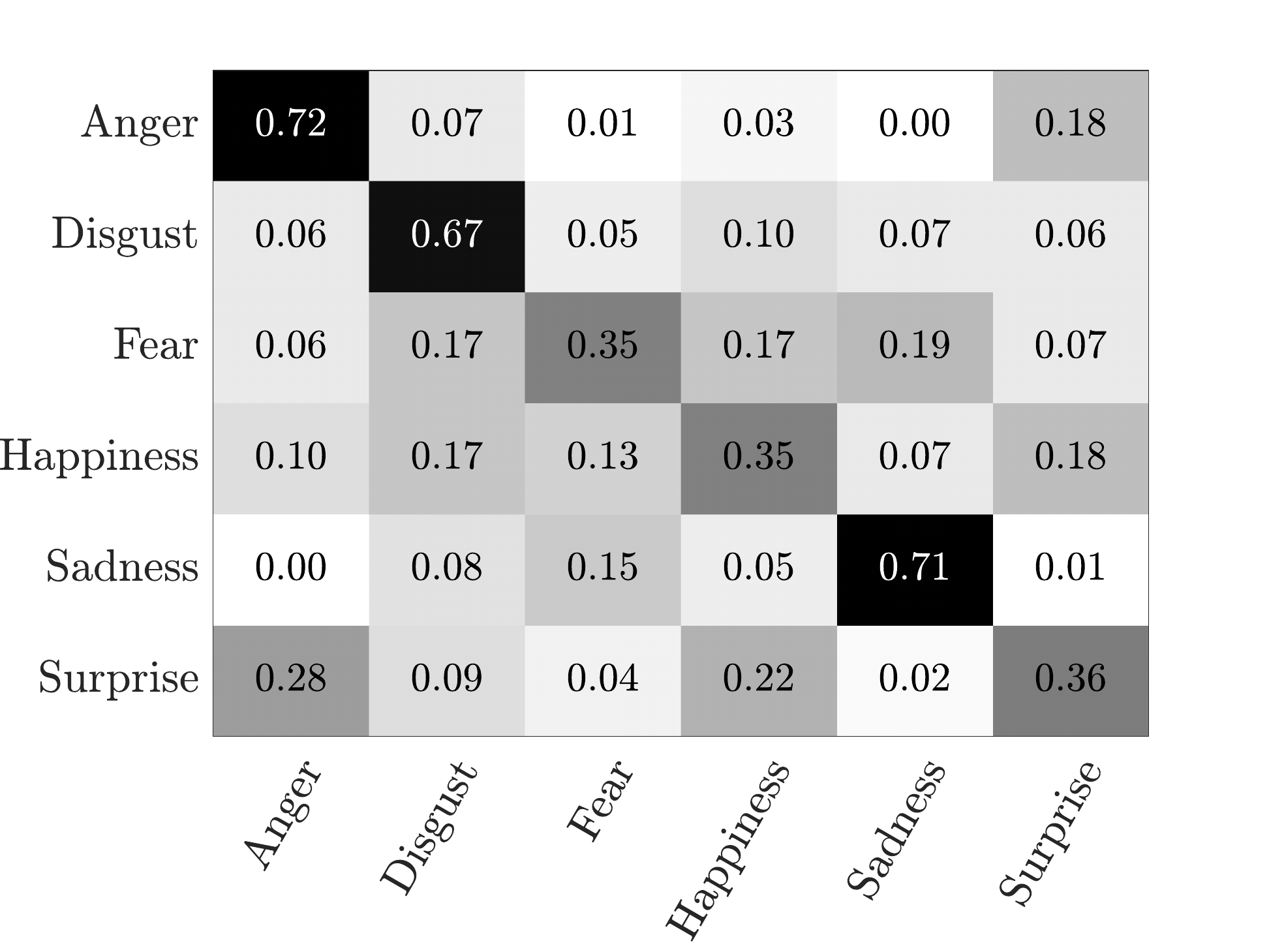}
   \caption{Normalised confusion matrices for the teacher model (left) and the student model (right) on the RML dataset (ground truth labels as rows, predictions as columns).}
\label{fig:confmats}
\end{figure}

\subsection{Discussion} \label{sec:discuss}
\noindent\textbf{Evaluation on external corpora:} 
Due to large variations in speech emotion corpora, speech emotion models work best if they are applied under circumstances that are
similar to the ones they were trained on~\cite{schuller2010cross}. For cross-corporal evaluation, most methods rely heavily on domain transfer learning or other
adaptation methods~\cite{zhang2011unsupervised,deng2014autoencoder,deng2014linked}. These works generally agree that cross-corpus evaluation works to a certain degree only if corpora
have similar contexts. We show in this work that the embeddings learnt on the \datasetName{} dataset can generalise to different corpora, even with differences in nature of the dataset (natural versus acted) and labelling scheme. While the performance of our student model falls short of the teacher model that was used to supervise it, we believe this represents a useful step towards the goal of learning useful speech emotion embeddings that work on multiple corpora without requiring speech annotation.\\
\noindent\textbf{Challenges associated with emotion distillation:}
One of the key challenges associated with the proposed method is to achieve a consistent, high quality supervisory signal by the teacher network during the distillation process.  Despite reaching state-of-the-art performance on the FERplus benchmark, we observe that the teacher is far from perfect on both the RML and eNTERFACE benchmarks.  In this work, we make two assumptions: the first is that distillation ensures that even when the teacher makes mistakes, the student can still benefit, provided that there is signal in the uncertainty of the predictions. The second is a broader assumption, namely that deep CNNs are highly effective at training on large, noisy datasets (this was recently explored in \cite{rolnick2017deep,mahajan2018exploring}, who showed that despite the presence of high label noise, very strong features can be learned on large datasets). To better understand how the knowledge of the teacher is propagated to the student, we provide confusion matrices for both models on the RML dataset in Figure \ref{fig:confmats}. We observe that the student exhibits reasonable performance, but makes more mistakes than the teacher for every emotion except sadness and anger.  There may be several reasons for this.  First, \datasetName{} used to perform the distillation may lack the distribution of emotions required for the student to fully capture the knowledge of the teacher.  Second, it has been observed that certain emotions are easier to detect from speech than faces, and vice versa \cite{busso2004analysis}, suggesting that the degree to which there is a redundant emotional signal across modalities may differ across emotions.   \\
\noindent\textbf{Limitations of using interview data:} Speech as a medium is intrinsically oriented towards
another person, and the natural contexts in which to study it are interpersonal. Interviews capture these interpersonal interactions well, and the videos we use exhibit real world noise. However, while the interviewees are not asked to act a specific emotion, i.e. it is a `natural' dataset, it is likely that celebrities do not act entirely naturally in interviews. Another drawback is the heavily unbalanced nature of the dataset where some emotions such as contempt and fear occur rarely. This is an unavoidable artefact of using real data. Several works have shown that the interpretation of certain emotions from facial expressions can be influenced to some extent by contextual clues such as body language \cite{aviezer2009not,hassin2013inherently}. Due to the \say{talking-heads} nature of the data, this kind of signal is typically not present in interview data, but could be incorporated as clues into the teacher network.\\
\noindent\textbf{Student Shortcuts:} The high capacity of neural networks can sometimes allow them to solve tasks by taking \say{shortcuts} by exploiting biases in the dataset \cite{doersch2015unsupervised}. One potential for such a bias in \datasetName{} is that interviewees may often exhibit consistent emotions which might allow the student to match the teacher's predictions by learning to recognise the identity, rather than the emotion of the speaker. As mentioned in Sec. \ref{s:experiments}, the performance of the student on the \texttt{heardVal} and \texttt{unheardVal} splits is similar ($0.71$ vs $0.69$ mean ROC AUC on a common set of emotions), providing some confidence that the student is not making significant use of identity as a shortcut signal. \\
\noindent\textbf{Extensions/Future Work:}
First, we note that our method can be applied as is to other mediums of
unlabelled speech, such as films or TV shows.  We hope to explore
unlabelled videos with a greater range of emotional diversity, which
may help to improve the quality of distillation and address some of
the challenges discussed above. Second, since the act of speaking may also
exert some influence on the facial expression of the speaker (for
example, the utterance of an \say{o} sound could be mistaken for
surprise)  we would also like to
explore the use of proximal {\em non-speech} facial expressions as a
supervisory signal in future work. Proximal supervision could also address the problem
noted in Section~\ref{sec:distillation}, that
speaking expressions can tend towards neutral. Finally, 
facial
expressions in  video can be learnt using 
self-supervision~\cite{wiles2018self}, and this offers an alternative to the strong supervision
used for the teacher in this paper.

%% file: conclusion.tex
\section{Conclusions} We have demonstrated the value of using a large
dataset of emotion unlabelled video for cross-modal transfer of
emotions from faces to speech. The benefit is evident in the results
-- the speech emotion model learned in this manner achieves reasonable
classification performance on standard benchmarks, with results far
above random. We also achieve state of the art performance on facial
emotion recognition on the FERPlus benchmark (supervised) and set 
benchmarks for cross-modal distillation methods for speech emotion
recognition on two standard datasets, RML and eNTERFACE.

The great advantage of this approach is that video data is almost
limitless, being freely available from YouTube and other sources.
Future work can now consider scaling up to larger unlabelled datasets,
where a fuller range of emotions should be available.

%% file: acknowledgements.tex
\noindent\textbf{Acknowledgements.}  The authors would like to thank
the anonymous reviewers, Almut Sophia Koepke and Judith Albanie for useful suggestions. 
We  gratefully acknowledge the support of EPSRC CDT AIMS grant
EP/L$015897$/$1$,  and the Programme Grant Seebibyte EP/M013774/1.